\documentclass[letterpaper]{article} 
\usepackage{aaai25}  
\usepackage{times}  
\usepackage{helvet}  
\usepackage{courier}  
\usepackage[hyphens]{url}  
\usepackage{graphicx} 
\usepackage{natbib}  
\usepackage{caption} 
\usepackage{algorithm}
\usepackage{algorithmic}

\usepackage{newfloat}
\usepackage{listings}
\DeclareCaptionStyle{ruled}{labelfont=normalfont,labelsep=colon,strut=off} 
\floatstyle{ruled}
\newfloat{listing}{tb}{lst}{}
\floatname{listing}{Listing}
\usepackage{booktabs}
\usepackage[most]{tcolorbox}
\usepackage{algorithm}
\usepackage{booktabs}
\usepackage{amssymb}
\usepackage{multirow}
\usepackage{booktabs}
\usepackage{rotating}
\usepackage{enumitem}
\usepackage{microtype}
\usepackage{mathtools}
\usepackage{amsmath}
\DeclareMathOperator*{\argmax}{arg\,max}

\usepackage{amsthm}
\usepackage{soul}
\usepackage{soul}

\newtheorem{mydef}{\textsc{{Definition}}}
\newtheorem*{prompt}{ChatGPT Prompt:}
\newtcolorbox{mytextbox}[1][]{%
  sharp corners,
  enhanced,
  colback=white,
  attach title to upper,
  #1
}

\newcommand{\mymethod}{\sc NoMatterXAI}
\title{{\mymethod}: Generating ``No Matter What'' Alterfactual Examples for Explaining Black-Box Text Classification Models}
\author{
    Tuc Nguyen\equalcontrib\textsuperscript{\rm 1},  James Michels\equalcontrib\textsuperscript{\rm 2}, Hua Shen\textsuperscript{\rm 3}, Thai Le\textsuperscript{\rm 1}\\
}
\affiliations{
    \textsuperscript{\rm 1} Department of Computer Science, Indiana University\\
    \textsuperscript{\rm 2} Department of Computer Science, University of Mississippi\\
    \textsuperscript{\rm 3} School of Information \& Computer Science and Engineering, University of Michigan—Ann Arbor\\
    tucnguye@iu.edu, jrmichel@go.olemiss.edu, huashen@umich.edu, tle@iu.edu
}

\usepackage{bibentry}

\usepackage{array}

\begin{document}

\maketitle

\begin{abstract}
In Explainable AI (XAI), counterfactual explanations (CEs) are a well-studied method to communicate feature relevance through contrastive reasoning of \textit{``what if''} to explain AI models' predictions. However, they only focus on important (i.e., relevant) features and largely disregard less important (i.e., irrelevant) ones. Such \textit{irrelevant features} can be crucial in many applications, especially when users need to ensure that an AI model's decisions are not affected or biased against specific attributes such as gender, race, religion, or political affiliation. 
To address this gap, the concept of \textit{alterfactual explanations} (AEs) has been proposed. AEs explore an alternative reality of \textit{``no matter what''}, where irrelevant features are substituted with alternative features (e.g., ``republicans''$\rightarrow$``democrats'') within the same attribute (e.g., ``politics'') while maintaining a similar prediction output. 
This serves to validate whether AI model predictions are influenced by the specified attributes.
Despite the promise of AEs, there is a lack of computational approaches to systematically generate them, particularly in the text domain, where creating AEs for AI text classifiers presents unique challenges.
This paper addresses this challenge by formulating AE generation as an optimization problem and introducing {\mymethod}, a novel algorithm that generates AEs for text classification tasks. Our approach achieves high fidelity of up to 95\% while preserving context similarity of over 90\% across multiple models and datasets.
A human study further validates the effectiveness of AEs in explaining AI text classifiers to end users. 
All codes will be publicly available.
\end{abstract}

%

\section{Introduction}
As AI advances, complex machine learning (ML) text classifiers have been developed to yield predictive performance competitively to that of humans for myriad tasks \cite{Pouyanfar2018-ef}. However, many of such models are so-called ``black-box'' models that are notorious for their lack of transparency. This may limit both the comprehension and societal acceptance of ML in critical fields, such as healthcare \cite{Tjoa2021-td}, finance \cite{Benhamou2021-cc}, and content moderation~\cite{kemp2021poll}. The field of Explainable Artificial Intelligence (XAI) \cite{Adadi2018-nl} aims to remedy this by explaining the factors at play in a model's predictions. 



A common paradigm found in XAI is counterfactual explanation (CE) \cite{Miller2019-eu} where an alternative reality is presented where minor alterations to input directly change an AI model's output applied to image, tabular, and text data classification problems \cite{Verma2020-vk, Garg2019-nd, Yang2020-ey}. CE follows the thought process of counterfactual thinking by asking ``\textbf{\textit{What if...?}}'', which is a common occurrence in the human psyche, through emotions such as regret, suspense, and relief \cite{Roese2009-ks}. CE is often delivered via natural language in the form of ``What if'' messages \cite{Le2020-ed,Hendricks2018-xw}. For example, a classifier that labels email messages as spam or ham could provide the text ``\textit{Had} the word `credit' and `money' is used twice in the message, it would have been classified as spam \textit{rather than} ham.'' \cite{Le2020-ed}.

\begin{figure}[tb!]
   \centering      
   \includegraphics[width=1\columnwidth]{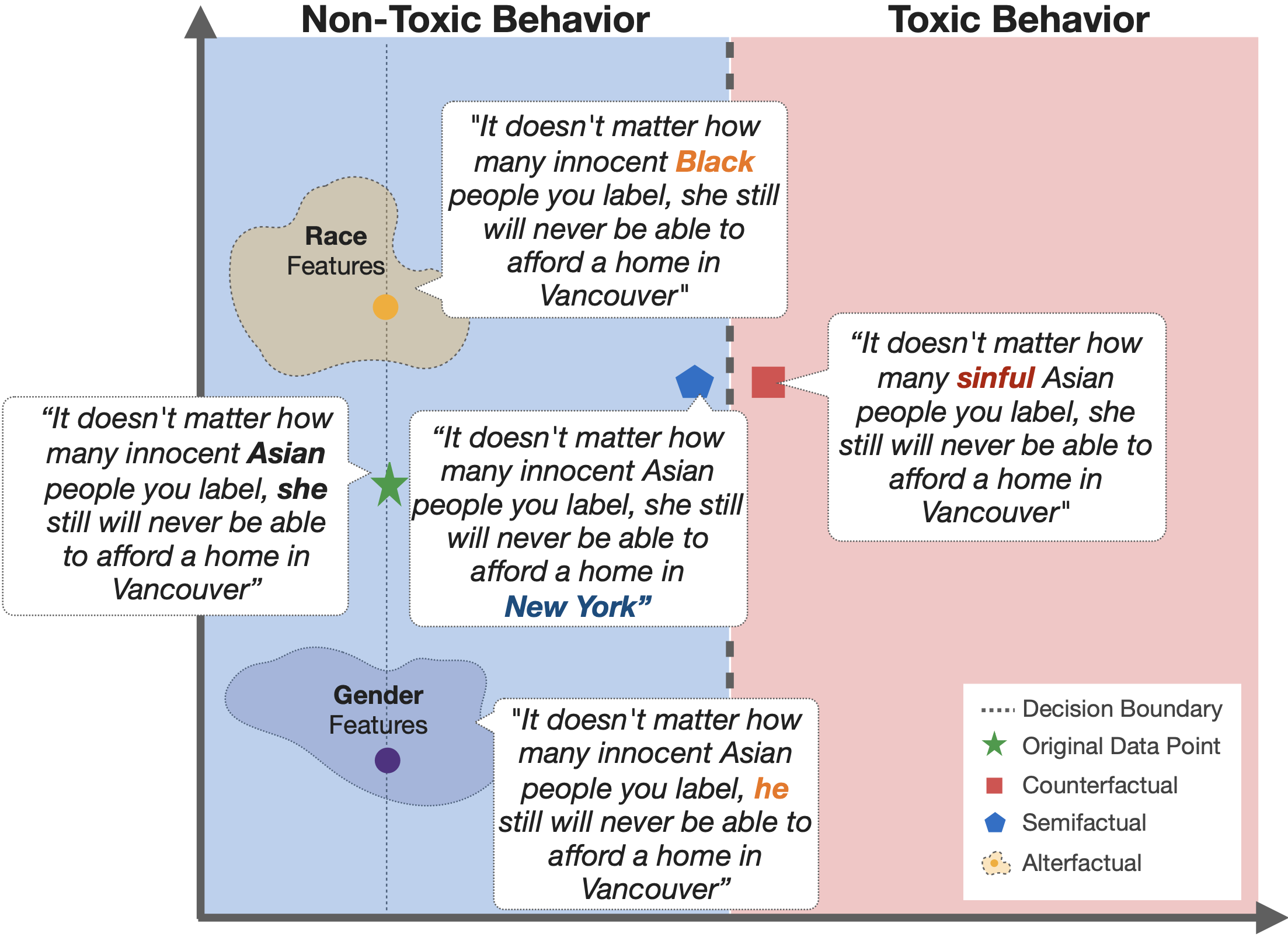}  
     \caption{A comparison of various AI explanation algorithms, including Counterfactual, Semifactual, and our proposed Alterfactals explanations. Alterfactual explanations aim to validate whether AI model predictions are influenced by specific attributes such as race or gender.
     }
      \label{fig:Alterfactuals}
      \vspace{-15pt}
\end{figure}
 
While CE is highly effective at providing intuitive reasoning to the users by emphasizing important features, it often neglects the role of less important ones in a text input, \textit{occluding information on what is indeed \underline{irrelevant} to a model's decision}. However, in many cases, irrelevant features are as important as relevant ones in explaining black-box predictions. For example, irrelevant features can help (1) contribute to the comprehensive understanding of a black-box model \cite{Mertes2022-xq} and (2) determine whether a model is biased against specific semantic features such as gender or race, which we cannot not fully understand with only CE.

A recent study posed a solution in the form of \textbf{\textit{alterfactual explanation}} (AE) \cite{Mertes2022-xq}. AEs embody the thought process of \textbf{\textit{``No matter what...''}} and present an alternative reality where a set of irrelevant features are \textit{significantly changed}, and yet the model's output remains the same. While~\cite {Mertes2022-xq} demonstrate that users view AEs equally favorably as counterfactual explanations, this was done with a hypothetical model for tabular data presented in the user study. The algorithmic generation of AEs for actual trained models is still needed. This can be achieved for tabular data by changing individual features significantly up to their domain ranges--e.g., alternating ``age'' of a patient from 0 to 100. However, in the NLP domain, textual features cannot be as directly altered due to their discrete nature, not to mention how to change a textual feature significantly but still maintain the reasonable semantic context of the original input--e.g., changing ``Republicans''${\rightarrow}$``Democrats'' as shown in Fig.~\ref{fig:Alterfactuals} is non-trivial. Thus, not only has the generation of AEs for text classifiers not been explored, but such a task also has its own unique challenges.

As a first step to exploring AEs for text classification tasks, this work investigates how to systematically generate alterfactual examples for text classifiers. We propose a framework, called {\mymethod}, that can \textit{significantly change different irrelevant features} of an input text to generate alterfactual for a target ML classification model. Our contributions are summarized as follows.

\begin{enumerate}[leftmargin=\dimexpr\parindent-0.2\labelwidth\relax,noitemsep]
  \item We elucidate a formal definition of AEs for text data. This definition is in an ideal theoretical form, and we explicate how it is translated to our solution.
  \item We introduce a novel algorithm {\mymethod}, which generates alterfactual variants of input texts, such that one or more irrelevant words are changed by opposite words selected via two strategies, ConceptNet and ChatGPT, while maintaining almost no noticeable changes in prediction probability and original context similarity.
  \item We conduct both automatic and human evaluations on four real-world text datasets, three text classifiers, achieving up to 95\% in effectiveness of generating AEs, showing that such AEs can support humans to accurately compare biases among different classification models.
\end{enumerate}

\begin{table}[tb]
\centering
\footnotesize
\begin{tabular}{p{2cm} p{5.7cm}}
\toprule
\multicolumn{1}{c}{\textbf{Type}} & \multicolumn{1}{c}{\textbf{Example}} \\ \cmidrule(lr){1-2}
Factual & \textbf{Since} your income is \$100K, you get the loan\\
Semifactuals & \textbf{Even if} your income is \$80K, you get the loan \\
Counterfactuals & \textbf{If} your income was \$1K lower, you \textbf{would had} not got the loan\\
\textbf{Alterfactuals} & \textbf{No matter what} your \textit{race} is, you would get the loan with your current income\\
\bottomrule
\end{tabular}
\caption{Examples of different types of explanations in a hypothetical scenario where an algorithm determines whether or not a person is approved for a loan based on their income.}
\label{table:examples}
\vspace{-10pt}
\end{table}

\section{Background and Motivation}
This section provides a summary of a variety of factual explanation examples applied to the NLP domain, including \textit{semifactuals}, \textit{counterfactuals}, and \textit{adversarials}. This will help distinguish the alterfactual from the rest (Table.~\ref{table:examples}).

\paragraph{Counterfactuals} are shown to be intuitive to humans by explaining \textit{``Why X, rather than Y''} for a model's decision such as \textit{``This email would be classified as ham rather than spam if there were 50\% less exclamation points''}~\cite{Le2020-ed}. Counterfactual explanations (CEs) are traditionally used in classification tasks \cite{Verma2020-vk} and recently information retrieval tasks~\cite{Kaffes2021-id,Agarwal2019-za,Tan2021-uf}. They tend to be minimal such that the input is perturbed \textit{as little as possible} to yield a contrasting output~\cite{Kenny2021-ci}. 

\paragraph{Semifactuals} explain \textit{``Even if X, still P.''}, or that an identical outcome occurs despite some \textit{noticeable} change in the input, providing explanation such as \textit{``This email is still spam even if it had 3 exclamation marks instead of 6''}. The exact definition varies, either as an input that is modified to be closer to the decision boundary~\cite{Kenny2021-ci} or others consider any input of the same class to be semifactual~\cite{Kenny2023-ae}.   

\paragraph{Adversarials} result from slight alterations to an input to fool an ML model's prediction. While closely related to CEs~\cite{Le2020-ed}, adversarial examples are different in their intent--i.e., confusing a model versus providing interpretability. They are similar to CEs, in that they are minimal changes with the intent of yielding a different classification. However, adversarial attacks are typically made to be not detectable by humans, while counterfactuals are to be detected and understood by humans. 

\paragraph{Alterfactuals} is a variant of semifactuals, proposed by Mertes et al. \cite{Mertes2022-xq} with the definition:
\begin{center}
\vspace{-5pt}
\fbox{\parbox[t]{0.95\linewidth}{
\begin{mydef}\label{def:def444} 
\normalfont{\bf Alterfactual Example in ML.}
Let denote $d$ be a distance metric on input space $X$, $d: X \times X \rightarrow R$. An alterfactual example of an example $x$ with a model $M$ is an altered version $x^* \in X$, that \textit{maximizes the distance} $d(x, x^*)$ with the distance to the decision boundary $B$ and the prediction of the model do not change--i.e., $d(x, B)\approx d(x^*, B)$ and $f(x){=}f(x^*)$.
\end{mydef}
}}
\vspace{-3pt}
\end{center}

\paragraph{Motivation.} While CEs present scenarios where negligible changes can alter an outcome, they focus less on explaining which features are \textit{irrelevant}. Since feature changes are minimal, statements may not encapsulate \textit{all} of the factors that play a role in a model's decision-making, which should include not only relevant but also irrelevant signals. AEs can identify irrelevant features by exaggerating their influence\cite{Mertes2022-xq}. This new perspective presents an intriguing option for explanation. However, Mertes' study solely focuses on measuring the effectiveness of AEs in explaining model behaviors to users. We want to examine how \textit{automatically generates AEs in practice} and aim to propose the first of its kind for text-domain.

\section{Problem Formulation}
Given a sentence $x$ and text classifier $M$, our goal is to generate new AE $x^*$, to provide interpretable information on irrelevant features of $x$ of the prediction $f(x)$. According to the \textit{Definition. 1}, we hope to generate AE $x^*$ is changing $x$ as much as possible, or:
\setlength{\abovedisplayskip}{-2pt}
\setlength{\belowdisplayskip}{0pt}
\begin{equation}
    \max\limits_{x^*}\;d(x, x^*)
    \label{eqn:max}
\end{equation}
\noindent Moreover, for $x^*$ an alterfactual example, it needs to maintain a similar distance to the decision boundary to the original predicted class and at the same time preserve the original prediction, or:
\setlength{\abovedisplayskip}{2pt}
\setlength{\belowdisplayskip}{5pt}
\begin{equation}
    \mathrm{argmax}(f(x^*)){=}\mathrm{argmax}(f(x)){\land}|f(x){-}f(x^*)|{\le}\delta,
    \label{eqn:constraint1}
\end{equation}
\noindent where $\delta$ is a small threshold constraining how much the original prediction probability can shift. 
However, without any additional constraint, $x^*$ might \textit{not} necessary preserve the same context of $x$ and can even result in meaningless sentences (e.g., ``today is monday''$\rightarrow$``today is school''). Thus, we want to perturb the original input $x$ (grey circle) to generate optimal $x^*$ that is \textit{also} \textit{furthest away} from $x$ and $x^*$ to be still within the context space of $x$, denoted as $\mathcal{S}_x$, or:
\setlength{\abovedisplayskip}{3pt}
\setlength{\belowdisplayskip}{2pt}
\begin{equation}
    x^* \in \mathcal{S}_x,
    \label{eqn:contextspace}
\end{equation}

It is, however, still non-trivial to systematically manipulate the whole sentence $x$ in the \textit{discrete} text space. Although we can manipulate $x$ via its embedding in the vector space, such manipulations may result in $x^*$ that is totally different from $x$ with many random changes that are no longer interpretable to the users. To tackle this, we can perturb $x$ through replacements of its individual words as similarly done in existing CE works. By replacing each word with its further possible alternative semantically--e.g., ``pretty''$\rightarrow$``ugly'', we hope to move the whole sentence $x$ as far as possible. We then opt for perturbing only \textit{irrelevant} features $x^*_\mathrm{ir}$ of $x^*$. Eq. \ref{eqn:max} becomes:
\setlength{\abovedisplayskip}{2pt}
\setlength{\belowdisplayskip}{-3pt}
\begin{equation}
    \max\limits_{x^*_\mathrm{ir}}\;d(x, x^*)
\end{equation}

Perturbation only irrelevant features $x_\mathrm{ir}$ of $x$ will also give us a more specific and intuitive ``no matter what'' explanation that we need such as in explanation: \textit{no matter what we change ``pretty'' (like to ``ugly'') in the sentence, the prediction would be still the same}. It will also help increase the chance $x^*$ is still parallel with the decision boundary as if we perturb not irrelevant but irrelevant or important features, it is more likely to change the prediction probability significantly. 

Still, we cannot replace $x_\mathrm{ir}$ with just any perturbation $x^*_\mathrm{ir}$. For example, good perturbations include antonyms--e.g., ``he''${\rightarrow}$``she'' as in {``\textit{no matter what} the gender of the person, the classifier still predicts hate-speech''}, or members of a distinct group--e.g., ``red'', ``blue'', ``green'' (colors), ``democrats'', ``republicans'' (political leaning) as in {``\textit{no matter what} the political leaning of the user, the classifier still predicts non-hate-speech'}. To enforce this constraint, we require that the replacement token needs to share the same \textit{semantic field}~\cite{jurafsky2000speech} with the original one, or:
\setlength{\abovedisplayskip}{3pt}
\setlength{\belowdisplayskip}{3pt}
\begin{equation}
    \mathrm{s}(x^*_\mathrm{ir}) = \mathrm{s}(x_\mathrm{ir})\;\forall\;x^*_\mathrm{ir}\in x^*,
    \label{eqn:semantic_field}
\end{equation}
\noindent where $x_\mathrm{ir}$ and $x^*_\mathrm{ir}$ denote arbitrary a pair of original and replacement word and $s(\cdot)$ queries the semantic field of a word. This constraint makes perturbations such as ``Monday''$\rightarrow$``cool'' in ``today is Monday and the weather is nice'' unfeasible because ``cool'' and ``Monday'' does not share the same semantic field, although ``cool'' is semantically far away from ``Monday'' and still somewhat preserves the original context. This results in the objective function below.

\begin{center}
\fbox{\parbox[t]{0.95\linewidth}{
\textbf{\textsc{Objective Function:}} For a given document $x$ with irrelevant features $x_{ir}$, text classifier $M$, and threshold hyperparameter $\delta$, our goal is to generate an alterfactual example $x^*$ of $x$ by solving the objective function:
\begin{equation}
\begin{aligned}
    \max\limits_{\{x^*_\mathrm{ir} \in x^*\}}\ \;\;&d(x, x^*) \;\;\mathrm{s.t.} \\
     \argmax(f(x^*)) &= \argmax(f(x)), \\ 
     d[f(x) - f(x^*)] &\le \delta, \\
    x^* &\in \mathcal{S}_x \\
    \mathrm{s}(x^*_\mathrm{ir}) &= \mathrm{s}(x_\mathrm{ir})\;\forall\;x^*_\mathrm{ir}\in x^*
\end{aligned}
\label{eqn:obj}
\end{equation}
}}
\end{center}

\section{{Proposed Method: \mymethod}}\label{sec:method}

\begin{figure}[tb!]
   \centering      
   \includegraphics[width=0.7\columnwidth]{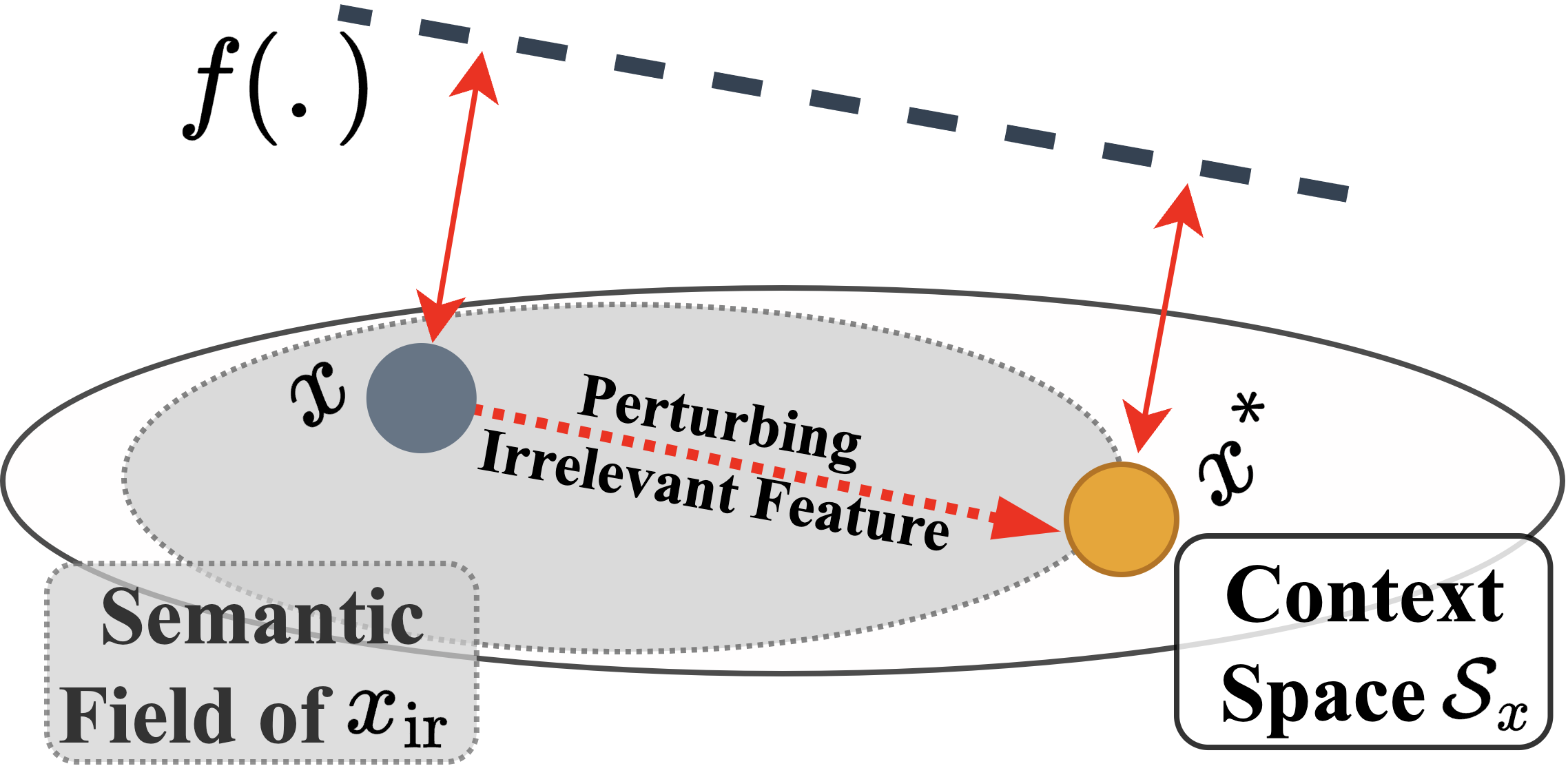} 
     \caption{AE generation of $x^*$ (orange circle) from $x$ (grey circle) by perturbing irrelevant features $x_\mathrm{ir}$ of $x$ within their semantic fields while still maintaining original context of $x$.}
      \label{fig:moving_context_space}
      \vspace{-15pt}
\end{figure}

To solve the objective function, we propose a novel greedy algorithm called {\mymethod}. Overall, {\mymethod} involves two steps. Given an input text $x$, it selects a maximum of $m$ words to perturb in order according to their importance to the prediction $f(x)$. Each word is greedily perturbed with its counterparts while ensuring all the constraints are satisfied. A detailed algorithm is described in the Alg.\ref{alg:attack}.

\subsubsection{Step 1: Irrelevant Feature Selection.}
Each feature of $x$ is ranked from lowest to highest predictive importance based on the probability drop in the original predicted class \textit{when they are individually removed} from $x$ (lines 2-5 in Alg.~\ref{alg:attack}). We prioritize perturbing features of lower importance--a.k.a., irrelevant features, first, since their perturbations are less likely to alter the model's prediction probability to the predicted class. Then, we iteratively transform one word at a time until we have checked a maximum of $m$ words (lines 6-16 in Alg.~\ref{alg:attack}). Hyper-parameter $m$ is set to ensure that (1) there are not too many perturbations in $x$ that could make the resulting AEs difficult to interpret and (2) reduce unnecessary runtime. 

\subsubsection{Step 2: Feature Perturbation with Opposite Word.}
We want to perturb the selected irrelevant features ``farthest away'' to their originals to move $x^*$ to right at the boundary of $\mathcal{S}_x$ as depicted in Fig.~\ref{fig:moving_context_space}. Moreover, such perturbations also need to share the same semantic field of the original token (Eq.~\ref{eqn:semantic_field}). We call these opposite words and adopt the definition of \textit{oppositeness} in terms of incompatibility in the linguistic literature--i.e., that is, for example, ``if a thing can be described by one of the members of an antonym pair, it is impossible for it to be described by the other''~\cite{keith2022knowing}. Such opposite words also often share the same semantic field of the original one~\cite{antonyms_field,jurafsky2000speech}. 

However, coming up with such perturbations is non-trivial as there is no clear quantitative measure for oppositeness for a word, and most of the relevant literature often desires semantically similar rather than opposite replacements such as in adversarial NLP. Even if we add noise to the original token's embedding to find replacements, it would require very different \textit{bound of noise} for different words to be still in the same semantic field--i.e., the grey region in Fig.~\ref{fig:moving_context_space} is dependent on $x_\mathrm{ir}$. For example, the $L_2$ distance between \textit{Glove} word embeddings~\cite{pennington2014glove} between ``pretty'' and ``ugly', ``Monday'' and ``Tuesday'', ``republicans'' and ``democrats'' are very different: 3.9, 0.4 and 1.3, respectively. 
Thus, adding a fixed amount much noise might end up in perturbations that are inappropriate.

Therefore, we adopt two different strategies that both leverage external knowledge to find opposite words for replacements: finding perturbations via ConceptNet database and large language model (LLM).

\begin{algorithm}[tb!]
\footnotesize
\caption{\textbf{AE Generation by {\mymethod}}}
\label{alg:attack}
\begin{algorithmic}[1]

\REQUIRE{Input sentence $x{=}\{ w_1,w_2,...,w_n\}$, target model $f(\cdot)$, sentence similarity threshold $\epsilon$, current perturbations $p_c$, current confidence score $c$, sentence similarity function $\mathrm{sim}(\cdot)$}. \\
\noindent \textbf{Output: }{AE $x^*$, confidence score post perturbation $c^*$}.
\STATE Initialize $x^*{\leftarrow}x$, $i{\leftarrow}0$, $p_c{\leftarrow}0$, $\delta{\leftarrow}0.05$, $c{\leftarrow}f(x)$
\FOR{each word $w_i{\in}x$}
\STATE Compute the importance score $I_{w_i}$.
\ENDFOR
\STATE Create a set $W$ of all words $w_i \in x$ sorted by the ascending order of their importance score $I_{w_i}$
\WHILE{$i \leq \operatorname{length}(W)$}
\STATE Find antonyms $\operatorname{\textit{a}}_i$ of $W[i]$ by ChatGPT or ConceptNet.
\STATE $x'\leftarrow$ Replace $W[i]$ with $\operatorname{\textit{a}}_i$ in $x^*$
\STATE $double\_negative\_check = \operatorname{Double\_Negative}(x')$
\IF {$double\_negative\_check$ == False}
\STATE $c'$ = $f(x')$
\STATE cond1 = $|c'$ - $c|\leq \delta$ AND $\mathrm{argmax}(c){==}\mathrm{argmax}(c')$
\STATE cond2 = $\mathrm{sim}(x,x'){\geq}\epsilon$
\IF{cond1 AND cond2}
    \STATE $c^* \leftarrow c'; x^* \leftarrow x'$
\ENDIF
\STATE $i \leftarrow i + 1$
\ENDIF
\ENDWHILE
\RETURN \texttt{($x^*, c^*$)}
\end{algorithmic}
\end{algorithm}

\paragraph{Opposite Words Selection via ConceptNet.}
The selected database for identifying antonymous words is the user-annotated knowledge base ConceptNet \cite{Speer2017-ik}. ConceptNet's word relations are notably annotated with numerical weightings through various sources. For a transformation of an input word, the following hierarchy of choices is used to identify opposite words (Table.~\ref{table:reasonableopps}).
\begin{itemize}[leftmargin=\dimexpr\parindent-0.2\labelwidth\relax,noitemsep]
    \item \textbf{Antonyms:} ConceptNet's API is called to check for words registered as the input word's antonym via the /r/Antonym relation, such that the weight of the relation is over $\omega_{t}$.
    \item \textbf{Distinct Items:} ConceptNet's API is called to check for words registered as members of a common set via the /r/DistinctFrom relation, such that something that is A is not B (e.g. red and blue), and that the weight of relation is over $\omega_{t}$. This ensures that choices of transformed words remain within common groups and can be adequately selected.
    \item \textbf{Hypernym's Hyponym:} We check for an umbrella term, referred to as a hypernym via ConceptNet's \textit{/r/IsA relation}, under which the input word belongs. For example, ``rose'', ``lilac'' and ``iris'' are all hyponyms of ``flowers''. If one is found, a query is made to identify members of the identified category that are not the input word, such that a member of the same category is to be selected. This is intended to identify words that are members of some overarching group, as similarly done in Distinct Items.
\end{itemize}


\paragraph{Opposite Words Selection via LLM.}
ChatGPT \cite{OpenAI2023-dv} estimates the likelihood of subsequent tokens in a text based on preceding words. We employ the inferred contextual understanding that ChatGPT can offer to identify antonyms. ChatGPT 3.5-Turbo is called for each input sentence and asked to provide one context-relevant antonym per word in the sentence such that the original sentence is still grammatically correct with the antonym replacement. Please refer to the Appendix for full details of the prompt.

\begin{table}[tb!]
\centering
\footnotesize
\begin{tabular}{p{1.5cm}p{6cm}}
\toprule
\textbf{Method} & \multicolumn{1}{c}{\textbf{Example}} \\ \cmidrule(lr){1-2}
Original & The children listened to \textcolor{black}{\textbf{jazz}} all day. \\
Antonym & The \textbf{\textcolor{black}{adults}} listened to jazz all day. \\
DistinctFrom & The children listened to jazz all \textbf{\textcolor{black}{month}}. \\
Hyponym & The children listened to \textbf{\textcolor{black}{rock}} all day. \\
\bottomrule
\end{tabular}
\caption{Examples of retrieved opposites from ConceptNet.}
\label{table:reasonableopps}
\vspace{-15pt}
\end{table}

\subsubsection{Avoiding Double Negatives.}
When words are changed for antonyms, some words have counterparts that are negative, such as "is" to "isn't". Multiple of these may cause double-negatives to arise in sentences, which may cause the user-interpreted meaning of the text to not significantly change. To address this,  we create a constraint to detect and reject potential double-negative sentences, unless the original text also featured a double-negative. This reduces potential confusing alterfactual texts to be returned to users. Of these replacements, we only keep those that do not create a double negative,  do not exchange a word for one that is a different part of speech (ex. noun$\rightarrow$verb), and that do not alter the model output confidence score $\delta$ beyond 5\%. The detailed algorithm is described in Alg.~\ref{double_negative_alg} (Appendix).

\newcolumntype{H}{>{\setbox0=\hbox\bgroup}c<{\egroup}@{}}
\renewcommand{\tabcolsep}{1.8pt}
\begin{table*}[tb!]
\footnotesize
\centering 
\begin{tabular}{ccccHHccc|ccHHccc|ccHHccc}
\toprule
& \multirow{2}{*}{\textbf{Method}} & \multicolumn{7}{c}{\textbf{DistilBERT}} & \multicolumn{7}{c}{\textbf{BERT}} & \multicolumn{7}{c}{\textbf{RoBERTa}}\\
\cmidrule(lr){3-9} \cmidrule(lr){10-16} \cmidrule(lr){17-23}
 &  & \multicolumn{1}{c}{\textbf{FID}$\uparrow$}  & \multicolumn{1}{c}{\textbf{AWP}$\uparrow$} & \multicolumn{1}{H}{\textbf{AVQ}$\downarrow$} &\multicolumn{1}{H}{\textbf{OPPL}$\downarrow$} &\multicolumn{1}{c}{\textbf{APPL}$\downarrow$} & \multicolumn{1}{c}{\textbf{SIM}$\uparrow$} & \multicolumn{1}{c}{\textbf{CON}$\downarrow$} &\multicolumn{1}{c}{\textbf{FID}$\uparrow$}  & \multicolumn{1}{c}{\textbf{AWP}$\uparrow$} & \multicolumn{1}{H}{\textbf{AVQ}$\downarrow$} &\multicolumn{1}{H}{\textbf{OPPL}$\downarrow$} &\multicolumn{1}{c}{\textbf{APPL}$\downarrow$} & \multicolumn{1}{c}{\textbf{SIM}$\uparrow$} & \multicolumn{1}{c}{\textbf{CON}$\downarrow$} & \multicolumn{1}{c}{\textbf{FID}$\uparrow$}  & \multicolumn{1}{c}{\textbf{AWP}$\uparrow$} & \multicolumn{1}{H}{\textbf{AVQ}$\downarrow$} &\multicolumn{1}{H}{\textbf{OPPL}$\downarrow$} &\multicolumn{1}{c}{\textbf{APPL}$\downarrow$} & \multicolumn{1}{c}{\textbf{SIM}$\uparrow$} & \multicolumn{1}{c}{\textbf{CON}$\downarrow$} \\
 \cmidrule(lr){1-23}
 \parbox[t]{4mm}{\multirow{5}{*}{\rotatebox[origin=c]{90}{\textbf{GB}}}} 
 & Feng et al & \textbf{95.56}  & \textbf{6.65} & 16.44 & 57.50 & 156.57 & 0.81 & 1.61 & \textbf{96.58} & \textbf{6.44} & 16.22 & 56.92 & 154.51 & 0.83 & 1.67 & \textbf{96.98} & \textbf{6.45} & 16.16 & 57.09 & 153.99 & 0.84 & 1.68\\
 & CNet-Single  & 77.78 & 1.00 & 8.90 & 54.83 & 86.41 & 0.86 & 1.43 &  79.44 & 1.00 & 8.82 & 54.74 & 86.15 & 0.86 & 1.31&  78.53  & 1.00 & 8.86 & 54.83 & 86.86 & 0.86 & 1.33\\
  & GPT-Single  & 70.83 & 1.00 & \textbf{8.25} & 56.99 & \textbf{83.90} & 0.86 & 1.36 & 69.49 & 1.00 & 8.15 & 57.01 & \textbf{83.58} & 0.86 & 1.21& 67.93  & 1.00 & 8.24 & 57.09 & \textbf{83.39} & 0.86 & 1.26\\
  & CNet-Multi  & 77.78 & 1.55 & 10.26 & 54.83 & 98.19 & 0.88 & \textbf{1.19}& 79.44  & 1.58 & 10.19 & 54.74 & 98.05 & 0.87 & 1.07& 78.55 & 1.57 & 10.22 & 54.83 & 94.29 & 0.87 & \textbf{1.10}\\
 & GPT-Multi  & 70.83 & 1.56 & 9.07 & 56.99 & 98.29 & \textbf{0.89} & 1.24 & 69.49 & 1.59 & 8.98 & 57.01 & 99.30 & \textbf{0.89} & \textbf{1.04} & 67.93 & 1.57 & 9.08 & 57.09 & 98.01 & \textbf{0.89} & 1.15\\
 \hline
\parbox[t]{4mm}{\multirow{5}{*}{\rotatebox[origin=c]{90}{\textbf{HS}}}} 
& Feng et al & \textbf{99.17} & \textbf{7.87} & 20.31 & 66.72 & 89.63 & 0.82 & 0.55 & \textbf{98.77} & \textbf{7.77} & 20.32 & 66.57 & 87.90 & 0.84 & 0.36& \textbf{99.18} & \textbf{7.84} & 20.19 & 66.89 & 90.93 & 0.85 & 0.36\\
& CNet-Single  & 92.26 & 1.00 & 12.30 & 66.11 & 76.18 & 0.87 & 0.47 &  92.21  & 1.00 & 12.29 & 66.15 & 76.49 & 0.87 & 0.36& 92.29 & 1.00 & 12.26 & 66.30 & \textbf{77.25} & 0.87 &0.25 \\
& GPT-Single  & 80.14 & 1.00 & \textbf{12.04} & 64.04 & \textbf{75.71} & \textbf{0.88} & 0.44& 79.32 & 1.00 & 12.01 & 66.93 & \textbf{75.94} & 0.88 & 0.32& 84.87 & 1.00 & 16.20 & 67.82 & 90.84 & \textbf{0.91} & 0.24\\
& CNet-Multi  & 92.26 & 2.34 & 14.47 & 66.11 & 88.48 & 0.84 & \textbf{0.33}& 92.21 & 2.40 & 14.44 & 66.15 & 90.09 & 0.84 & \textbf{0.27} & 92.29 &2.46 & 14.41& 66.30 & 89.62 & 0.85 & \textbf{0.18} \\
& GPT-Multi  & 80.28 & 2.24 & 13.56 & 67.04 & 87.06 & 0.87 & 0.37 & 79.32 & 1.21 & 12.24 & 66.93 & 87.34 & \textbf{0.88} & 0.31& 84.87  &3.47 & 18.60 & 68.72 & 98.93 & 0.87 & 0.20\\
 \hline
\parbox[t]{4mm}{\multirow{5}{*}{\rotatebox[origin=c]{90}{\textbf{JIG}}}} 
& Feng et al & \textbf{97.29} & \textbf{26.82} & 73.42 & 70.07 & 121.00 & 0.81 & 0.83 & \textbf{97.76} & \textbf{27.77} & 71.34 & 71.00 & 124.6 & 0.85 & 0.76& \textbf{96.65} & \textbf{26.27} & 72.05 & 68.67 & 149.87 & 0.84 & 0.58\\
& CNet-Single  & 89.79 & 1.00 & \textbf{28.82} & 68.26 & 76.81 & \textbf{0.92} & 1.37& 89.99 & 1.00 & 28.62 & 69.43 & \textbf{78.18} & 0.92 & 1.59& 90.10 & 1.00 & 29.31& 67.31 & 75.68 & 0.92 & 0.88\\
& GPT-Single  & 82.45  & 1.00 & \textbf{28.61} & 68.19 & \textbf{75.02} & \textbf{0.92} & 1.35& 83.35 &1.00  & 45.35& 69.58 & 77.25  & \textbf{0.93} & 1.59& 80.38 & 1.00 & 46.31 & 68.3 & \textbf{75.17} & \textbf{0.93}& 0.86\\
& CNet-Multi & 89.83 & 3.91 & 41.33 & 68.26 & 105.85 & 0.88 & \textbf{0.52}& 89.99 & 3.62 & 41.02& 70.04 & 104.58 & 0.88 & 0.62& 90.10 & 5.13 & 41.97 & 68.09 & 116.47 & 0.88 & \textbf{0.27}\\
& GPT-Multi  & 82.52  & 3.93 & 38.39 & 68.19 & 98.19 & 0.89 & 0.64 & 83.35  & 6.51 & 62.26 &  70.04 & 106.39 & 0.90 & \textbf{0.61}& 80.38 & 10.40 & 62.63& 68.09 & 114.99 & \textbf{0.93} & \textbf{0.27}\\
 \hline
\parbox[t]{4mm}{\multirow{5}{*}{\rotatebox[origin=c]{90}{\textbf{EMO}}}} 
& Feng et al & \textbf{98.65} & \textbf{12.80} & && 254.32 & 0.81 & 0.42 & \textbf{99.78} & \textbf{11.60} & & & 280.84 & 0.81 & 0.68 & \textbf{99.89} & \textbf{11.14}  & & & 349.74 & 0.80 & 0.43 \\
& CNet-Single & 95.68  & 1.00 & \textbf{12.52} & 68.42 & \textbf{92.17} & \textbf{0.89} & 0.24 & 94.83 & 1.00 & 12.71 & 68.05 & \textbf{92.57} & \textbf{0.89} & 0.57 & 95.20 & 1.00  & 12.65 & 68.8 & \textbf{93.02} & 0.89 & \textbf{0.30} \\
& CNet-Multi & 95.68 & 3.11 & 15.99 & 68.42 & 158.70 & 0.84 & 0.19 & 94.83 & 2.66 & 16.42 & 68.25 & 141.56 & 0.86 & 0.48 & 95.20 &2.96 & 16.10& 68.64 & 150.6 & 0.85 & 0.23 \\
& GPT-Single & 86.93  & 1.00 & 13.02 & 68.64 & 95.01 & 0.88 & 0.20& 90.31 & 1.00 & 22.81 & 68.49 & 94.06 & 0.87 & 0.54 & 89.65  & 1.00 & 22.64 & 68.5 & 96.42 & \textbf{0.90} &  0.31\\
& GPT-Multi  & 86.93  & 3.10 & 15.56 & 68.64 & 146.34 & 0.85 & \textbf{0.16}& 90.31 &4.26 & 27.37 & 68.25 & 153.09 & 0.84 & \textbf{0.46} &  89.65 & 4.60 & 26.61 & 68.64 & 165.27 & 0.84 & 2.52\\
 \hline
 \end{tabular}
\caption{Summary of quantitative performance comparisons of {\mymethod}.}
\label{full_results}
\end{table*}

\section{Experiment Settings}
This section shows a comprehensive evaluation of {\mymethod} with different settings and baselines.

\paragraph{Datasets and Models.}
We use datasets of varied tasks, including gender bias(GB)~\cite{Dinan2020-oo}, hate speech classification (HS)\cite{Davidson2017-xu}, emotion classification (EMO) \cite{Saravia2018-xm}, and the toxicity detection in social comments (JIG) \footnote{\url{https://huggingface.co/datasets/james-burton/jigsaw_unintended_bias100K}}. They vary in average sentence length (9.3, 13.72, 43.38, 19.15 tokens) and number of labels (2,2,2,6). Each dataset is split into 80\% training and 20\% test splits, and we use the training set to train three target models, namely DistilBERT~\cite{Sanh2019-ju}, BERT~\cite{BERT} and RoBERTa~\cite{RoBERTa}.
Please refer to Table.~\ref{TableModels} (Appendix) for more details.

\renewcommand{\tabcolsep}{1.8pt}
\begin{center}
\begin{table}[tb!]
\footnotesize
\centering 
\begin{tabular}{p{4cm}p{4cm}}
\hline
\multicolumn{1}{c}{\textbf{Original $\longrightarrow$}} & \multicolumn{1}{c}{\textbf{Alterfactual Example}}\\
\toprule
Your comment makes no sense and is \textcolor{red}{incoherent} & Your comment makes no sense and is \textcolor{red}{coherent}\\
\cmidrule(lr){1-2}
Impossible to \textcolor{red}{understand} the stupidity of someone [...] & Impossible to \textcolor{red}{misunderstand} the stupidity of someone\\
\cmidrule(lr){1-2}
Mulcair's comment was \textcolor{red}{silly}, to say that the woman was 'illegally' \textcolor{blue}{refused} entry to the US. Obviously it is perfectly \textcolor{purple}{legal} for the US [...]. My guess is that the refusal was based on \textcolor{green}{her} purported engagement to a US citizen [...] situation carefully. & Mulcair's comment was \textcolor{red}{mature},  to say that the woman was 'illegally' \textcolor{blue}{approved} entry to the US. Obviously it is perfectly \textcolor{purple}{illegal} for the US [...]. My guess is that the refusal was based on \textcolor{green}{his} purported engagement to a US citizen  [...] carefully. \\
\bottomrule
\end{tabular}
\caption{Examples of AEs generated by {\mymethod}.}
\label{alterfactual_examples}
\vspace{-10pt}
\end{table}
\end{center}

\paragraph{Evaluation Metrics.}
We report the following metrics: Fidelity (FID$\uparrow$), or the percentage of texts that we are able to generate an AE; Runtime (Time$\downarrow$); Average Words Perturbed (AWP$\downarrow$); Average Queries (AVQ$\downarrow$) or an average number of queries made to target models; Altered Perplexity (APPL$\downarrow$), or the naturalness of $x$ and $x^*$ captured via GPT2-Large as a proxy~\cite{Radford2019-jq}; semantic similarity through the USE Encoder~\cite{Cer2018-vh} (SIM$\uparrow$); and the models' average confidence shift (in \%) after perturbations (CON$\downarrow$). 

\paragraph{Implementation Details.}
We select our confidence threshold $\delta{\leftarrow}0.05$ to allow the model output to only shift at most 5\% in confidence. Constraint Eq. (\ref{eqn:contextspace}) is satisfied by setting a minimum context similarity threshold $\epsilon{=}0.8$ via USE Encoder~\cite{Cer2018-vh}. We constrain {\mymethod}'s perturbations by preventing repeat perturbations and disregarding a list of stopwords. During perturbation, a word is not altered if either ConceptNet or GPT fails to return an option. Please refer to the Appendix for full details.

\paragraph{Baseline.} We evaluated two variants of {\mymethod}, one uses ConceptNet (CNet) and another uses ChatGPT LLM (GPT) for looking up replacement candidates. We also test {\mymethod} when perturbing only one word (denoted by \textit{``-Single''} suffix) and when perturbing as many words as we can (denoted by \textit{``-Multi''} suffix). Since there is no existing method that specifically generates AEs, we adopt \citep{Feng2018-af}, a method that iteratively removes the least important word from the input as an additional baseline.

\section{Results}\label{sec:results}
Table \ref{alterfactual_examples} depicts a few AEs synthesized by {\mymethod}. We describe in detail evaluation results on different computational aspects below, followed by a user-study experiment that evaluates the explainability of the generated AEs in practice with human subjects.

\paragraph{Generation Success Rate--i.e., Fidelity (FID$\uparrow$).} 
Being the first of its kind, {\mymethod} is able to find AEs around 70\% up to 95\% of the time. The baseline ~\cite{Feng2018-af} has a better chance of finding AEs by iteratively removing a set of least important words (Table.~\ref{full_results}), it totally discards the original contextual meaning of the sentence. This happens because \textit{deleting too many words would cause the resulting sentences to lose both semantic coherence and grammatical correctness}. As a result,~\cite{Feng2018-af} baseline results in a significantly higher (undesirable) perplexity on the perturbed samples and much lower reports on context preservation compared to {\mymethod}.

\paragraph{Context Preservation--i.e., Context Similarity (SIM$\uparrow$).} Baseline \cite{Feng2018-af} consistently ranks lower in context preservation to {\mymethod} (Table.~\ref{full_results}). This suggests that simply removing words fails to preserve the meaning of the original sentence. In contrast, using LLM like ChatGPT to generate replacement candidates yields the highest similarity in most cases. This happens because LLMs are well-designed to capture semantic meaning in natural language from vast amounts of data~\cite{chang2024survey}.

\paragraph{Changes in Prediction Probability (CON$\downarrow$).} Due to the constraints of the search condition, we observe that the alterfactual examples generated by {\mymethod} do not move significantly away from the original predicted class, as reflected by the near-zero average changes in prediction probabilities of 0.73\%, 0.77\%, and 0.71\% for DistilBERT, RoBERTa, and BERT, respectively. This indicates that {\mymethod} can produce alterfactual examples that diverge from the input while remaining aligned with the original model's decision boundary.

\begin{figure}[tb!]
   \centering      
   \includegraphics[width=1\columnwidth]{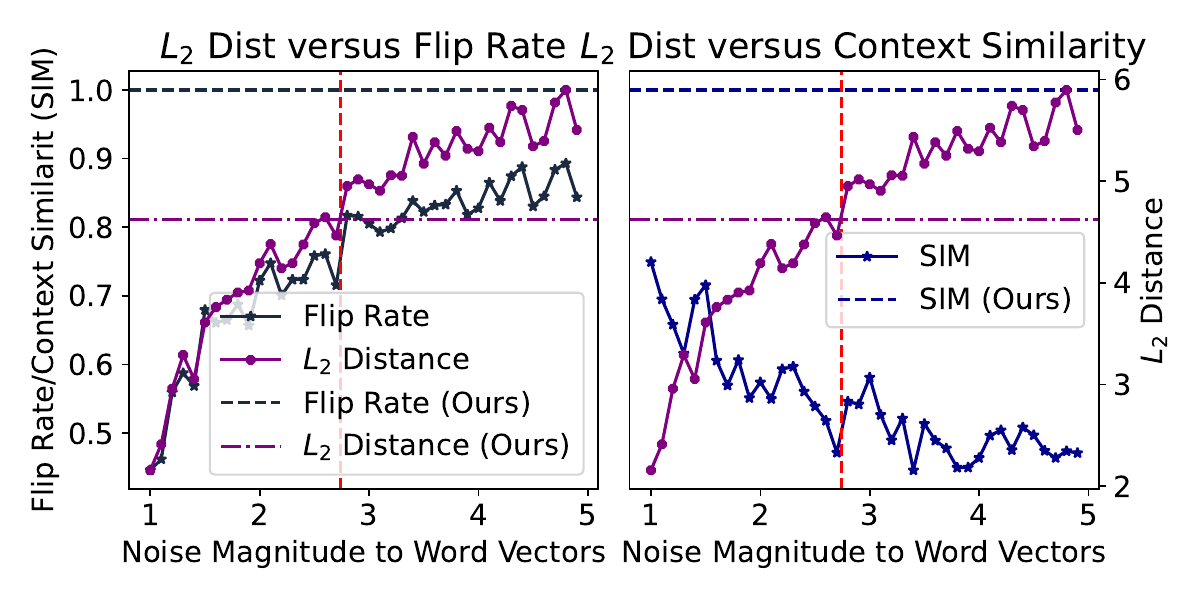} 
     \caption{Trade-off between $L_2$ distance between word embeddings of original and perturbed token versus Flip Rate--i.e., the chance of perturbed token converting to new word, and context similarity (SIM) on DistilBERT with JIG dataset.}
      \label{fig:tradeoff}
      \vspace{-10pt}
\end{figure}

\paragraph{Comparison with Alternative Perturbation Strategy.} We compare the use of ConceptNet against an alternative strategy of perturbation by adding noise to word embeddings as analyzed in $\S$\ref{sec:method} and Fig.~\ref{fig:moving_context_space}. To do this, we add Gaussian noises of incrementally increasing in magnitude to the embeddings of the original tokens and check (1) whether the resulting embeddings actually convert to a new token (Flip Rate) and whether the resulting sentences preserve the context similarity (SIM$\uparrow$). Fig. \ref{fig:tradeoff} shows that {\mymethod} is able to select suitable opposite words while maximizing SIM with much fewer changes in embedding space measured by $L_2$. This shows that such an alternative strategy will not work in practice as it significantly drifts from the original context as bigger noise is added to ensure a high Flip Rate. ConceptNet is more suitable for finding opposite words, which might not be systematically quantifiable in the embedding space.

\begin{figure}[tb!]
   \centering      
   \includegraphics[scale=0.5]{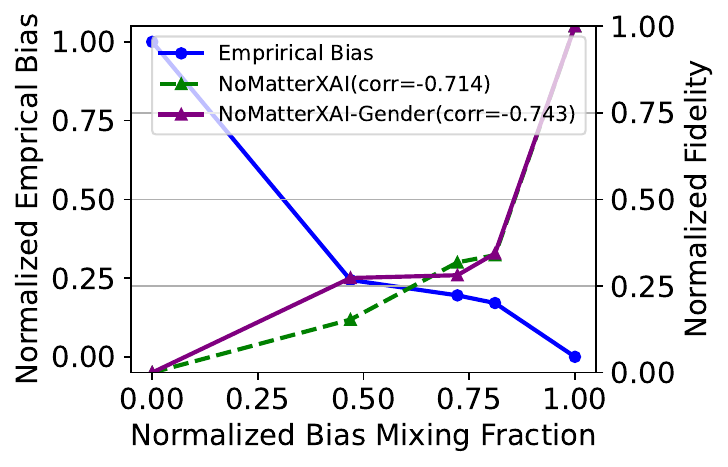}   
 \caption{{\mymethod}'s fidelity has a strong negative correlation (correlation coefficient \textit{corr}${\le}$-0.7) with the empirical gender bias evaluation (scores are normalized to [0,1])}
 \label{fig:Bias}
\end{figure}

\paragraph{Correlation with Model Bias Detection.} Since AEs emphasize irrelevant features, a model that is highly biased against gender should result in almost no AEs--i.e., near zero fidelity when we only perturb identity words--e.g., ``she'', ``he''. Similarly, an unbiased model should result in high fidelity. To further evaluate the generated AEs' qualities, we measure how well {\mymethod}'s fidelity correlates with automated bias detection metrics, especially when we target identity words to perturb. Fig.~\ref{fig:Bias} confirms the quality of {\mymethod}. This also shows the potential utility of {\mymethod} in approximating bias levels of text classifiers.

\section{User Study Experiment}

In this section, we evaluate the effectiveness of AEs with end users recruited from Amazon Mechanical Turk (MTurk). Our user study aims to answer the question: \textit{Can AEs be useful for humans to judge the model fairness?} We elaborate on our hypothesis, study details, and results below.

\paragraph{Hypothesis.} We evaluate whether AEs generated by {\mymethod} can inform the users about the relative bias rankings among three AI models of different empirical bias levels borrowed from $\S\ref{sec:results}$ (A-17.1\%, D-5.5\%, and E-0.7\%). Such rankings are significantly useful in practice to decide which AI models should be prioritized for deployment. Particularly, we define three alternative hypotheses $\mathcal{H}_a$ (Table.~\ref{tab:userstudy}) to validate whether or not college-level users can correctly identify three \textit{pair-wise} rankings better than a random guess by using explanations generated by {\mymethod}.

\begin{table}[tb!]
    \centering
    \footnotesize
    \begin{tabular}{ccccc}
        \toprule
        $\mathcal{H}_a$ & \textbf{Alternative Hypothesis} & \textbf{df} & \textbf{t-test} & \textbf{p-value}\\
        \cmidrule(lr){1-5}
        $\mathcal{H}_1$ & Correct Ranking: A$>$E ($\Delta{=}16.4\%$) & 45 & 2.01 & 0.026$^*$\\
        $\mathcal{H}_2$ & Correct Ranking: A$>$D ($\Delta{=}11.6\%$) & 36  & 2.94 & 0.003$^{**}$\\
        $\mathcal{H}_3$ & Correct Ranking: D$>$E ($\Delta{=}4.8\%$) & 39 & 8.075 & 4.5e-10$^{**}$\\
        \bottomrule
        \multicolumn{5}{l}{(*), (**) statistical significance with $\alpha{=}0.05$ and $\alpha{=}0.01$}
    \end{tabular}
    \caption{User study experiment results with different $\mathcal{H}_a$ of different gaps $\Delta$ in empirical bias scores.}
    \label{tab:userstudy}
    \vspace{-10pt}
\end{table}

\paragraph{Study Design.} Whether or not a model is biased \textit{cannot} be quantified with individual prediction instances. To evaluate such property, we perturb all gender words on 500 test examples curated from the JIG dataset to generate AEs and use them to curate a text explaining this global behavior along \textit{``No matter what we changed the genders mentioned in the input texts (like male${\rightarrow}$female, she${\rightarrow}$he, woman${\rightarrow}$man, etc.), the computer system's decisions remained the same for 1.8\% of the time''}. We present such an explanation for each of the two models--e.g., A\&D, A\&E, etc., and ask the participants to rank \textit{which model is less biased towards gender?}. We also include a simple definition of bias in AI models in the instruction. 
Please refer to the Appendix for more details.

\paragraph{Participant Recruitment and Quality Assurance.} We recruited adult (${>}18$ years old) participants from the USA on MTurk without assuming any knowledge of AI or ML. We pay each completed response US\$0.50 for roughly 2 minutes of work, resulting in \$12/hour average wage. We employ a three-layer quality assurance procedure. First, we utilize worker tags provided by MTurk to only select subjects having done at least 5,000 tasks with over ${\geq}98\%$ acceptance rate and completing U.S. Bachelor's degree. Second, we deploy a trivial attention check question to make sure the workers read and understand the instructions. Third, we provided incentives to the workers as an additional bonus payment of US\$0.50 for every correct answer to encourage their attention to the task. We also record the time each worker spends on the study to filter out obvious low-quality responses.

\paragraph{Results.} We collected responses from a total of 149 workers and discarded data from 29 workers due to (i) low attention time ($\leq$10 seconds) and/or (ii) incorrect answers to the attention check question. It is statistically significant to reject the null hypothesis in all cases using a one-sample t-test (ranking accuracy$>$0.5) (Table.~\ref{tab:userstudy}). This shows that explanations synthesized from AEs can effectively support the users to effectively compare the models' biases. On average, we also observe that workers who passed the attention question were both more confident (p-values${<}$0.05, except $\mathcal{H}_1$) and accurate (p-values${<}$0.05) at answering the ranking question. This shows that a minimal understanding of bias in AI models is a prerequisite for our task and the inclusion of such attention-check questions was crucial. 


\setlength{\belowcaptionskip}{-10pt}
\begin{figure}[tb!]
   \centering      
   \includegraphics[scale=0.16]{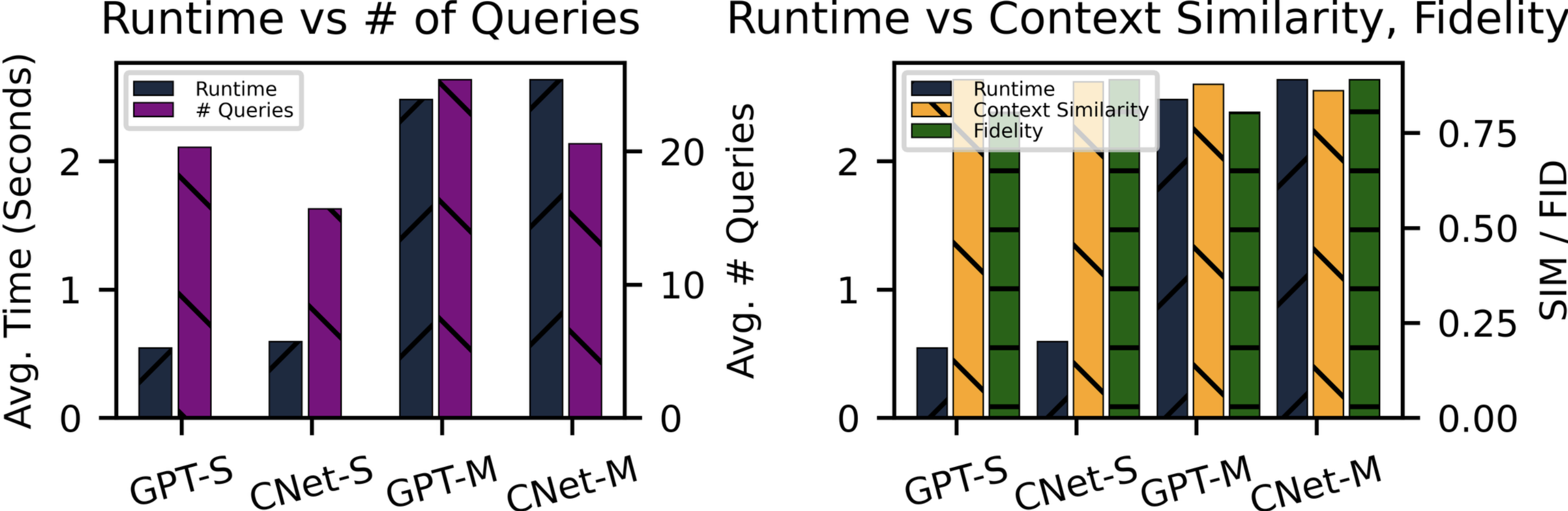}   
 \caption{Trade-off between runtime, number of queries, fidelity and context similarity per input, and number of model queries averaged across all datasets and target models.}
 \label{fig:avg_runtime}
\end{figure}

\vspace{-0.2cm}
\section{Discussion}
\paragraph{Computational Complexity Trade-Off.}
In this section, we analyze the time complexity of {\mymethod} algorithm (Alg. ~\ref{alg:attack}) on each input example. Computing the important scores takes $O(kV)$, where $V$ is the time complexity of a forward pass or query to the target classifier, and $k$ is the number of words in the original sentence. Sorting the list of $k$ importance scores takes $O(k{\operatorname{log}}k)$ with QuickSort. Finding opposite words and checking for the constraints takes $O(kV)$. To sum up, the overall time complexity of {\mymethod} to generate an AE for one instance is $O(k{\operatorname{log}}k{+}kV)$. 

Fig. \ref{fig:avg_runtime} confirms our analysis, showing that runtime highly is correlated with the number of queries to the target models. Although perturbing multiple words helps increase fidelity, such an effect is negligible compared to a significant increase in runtime, given that the context similarity remains more or less the same. This shows that one might consider generating AEs with only a few targeted words (like gender or race identities) in specific applications such as bias evaluation.

\begin{figure}[tb!]
\centering      
\includegraphics[scale=0.43]{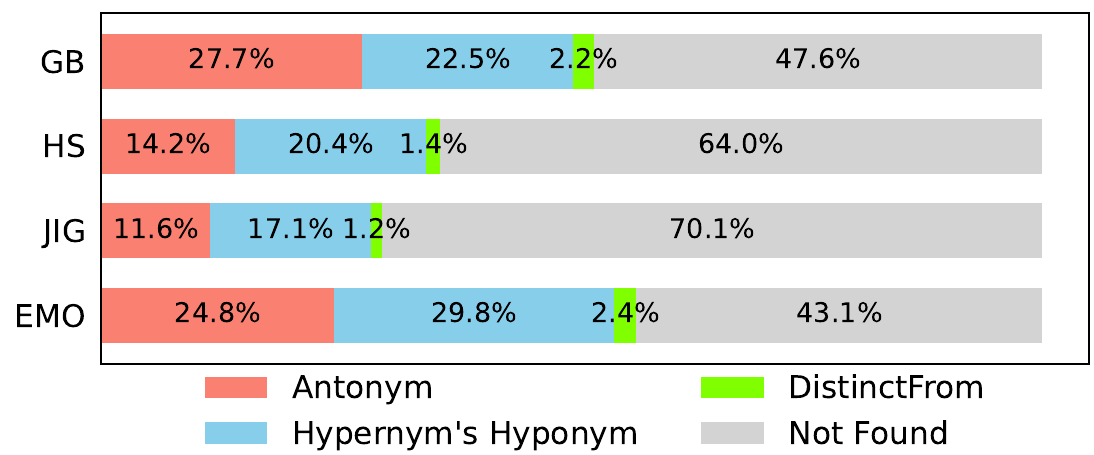}   
\caption{Categories of words returned from ConceptNet.}
\label{fig:conceptnet}
\end{figure}

\paragraph{Limitations of Perturbations with ConceptNet and ChatGPT.}  
ConceptNet~\cite{Speer2017-ik} is tied to the limited contents of its database. Some antonyms such as ``glow'' to ``dim'', are not present in the database at the time of writing. Additionally, a significant number of query calls yielded no result (Fig.~\ref{fig:conceptnet}). 
From our analysis of ConceptNet versus an alternative strategy in $\S\ref{sec:results}$, we once again emphasize that quantitatively finding opposite words is very challenging. 

While ChatGPT 3.5 is effective at generating opposite words most of the time, hallucinations do occur--e.g., replacing queried words with ``antonym'', although only on rare occasions. ChatGPT would occasionally return the dictionary \textit{not} in the requested JSON format (at the time of the experiment). This shows to have been addressed by the recent rollout of the ``structured output'' feature from OpenAI.

\paragraph{Other Limitations.}
A limitation with generating AEs as compared to existing explanations is the increased runtime. CEs are minimal in nature such that as few words as possible are perturbed, as compared to {\mymethod}, which aims to perturb as many words as possible. 
It is unclear if this will reduce the incentive to use AEs as compared to counterfactual examples, although their efficacy was shown to be similar in a previous evaluation with humans~\cite{Mertes2022-xq}.

\vspace{-0.2cm}
\section{Conclusion and Future Work}
In this paper, we extend the theoretical definition of alterfactuals \cite{Mertes2022-xq} to propose {\mymethod}, an automatic greedy-based mechanism that is able to generate alterfactual examples up to 95\% of the time to explain text classifiers. Through a human study, AEs generated by {\mymethod} show to help synthesize ``no matter what'' XAI texts to convey to users the irrelevancy in predictive features and reveal comparative bias behaviors among several target models. Future works include improving the knowledge base of database-oriented methods like ConceptNet or improving prompts for LLM-based opposite words identification.

\newpage
\clearpage

\bibliography{aaai25}

\appendix
\newpage

\section{LLM Prompt}
LLMs like ChatGPT \citep{OpenAI2023-dv} predict the likelihood of subsequent tokens in a text-based on preceding words. We utilize the contextual understanding provided by LLMs to identify antonyms. For each input sentence, we invoke ChatGPT 3.5-Turbo with a prompt optimized to convey the most relevant information concisely.

\begin{prompt}
"Job: output context-relevant antonyms for each word in a sentence. Output: JSON table with one row per word, each word is followed by ONE context-relevant antonym. Each antonym should be a single word. The original sentence should be grammatically correct when the antonym is swapped in. No titles, just "Word:Antonym". Words with no antonym should pair with '-'."
\end{prompt}

After the prompt is invoked, a dictionary is returned, which is then used for word perturbation. If a '-' token is returned, that word is excluded from the perturbation process.

\section{Double Negative Detection}
\label{double_negative_detection}

\begin{algorithm}[]
\caption{Double Negation Detection}
\label{double_negative_alg}
\begin{algorithmic}[1]
\STATE \textbf{Input:} A sentence $x$, predicted negativity threshold $n_t$, window size $s$, negativity prediction model $N$.
\STATE \textbf{Output:} Sentence $x$ is have double negation or not.
\STATE \textit{Initialize} list of negative items $L \gets \varnothing$
\IF{$len(x) == 0$}
    \STATE return $False$
\ENDIF
\WHILE{$len(x) > 0$}
    \STATE $w_i, S_n = N(x)$ \COMMENT{find negative word(s) and it's probability}
    \IF{$S_n \leq n_t$}
        \STATE Append $w_i$ to list $L$
    \ELSIF{$S_n < 1E-7$}
        \STATE break
    \ENDIF
    \FOR{each negative word $L_i$ in $L$}
    \STATE Remove word $L_i$ from sentence $x$
    \ENDFOR
\ENDWHILE
\IF{any negative word $L_i$ in list $L$ is within $s$ window size of another negative word in the original text:}
    \STATE return $True$
\ELSE 
    \STATE return $False$
\ENDIF
\end{algorithmic}
\end{algorithm}
Alg.~\ref{double_negative_alg} illustrates the steps to evaluate the number of negative examples in a sentence. A question-answering model is employed to detect negative words in the input sentence \footnote{\url{https://huggingface.co/Ching/negation_detector}}. The text is iteratively evaluated to identify and remove negative words. Once all negative words are isolated, the distance between each pair is checked. If the distance falls below a window size $w$, the sentence is considered to contain a double negative. The process runs Alg.~\ref{double_negative_alg} on each sentence of the original and perturbed texts. If any sentence that originally did not contain a double negative is perturbed to include one, then the specific perturbed text is rejected as a possible alterfactual.

To determine sufficient values for $n_t$ and $w$, Alg.~\ref{double_negative_alg} is evaluated on a small dataset generated by ChatGPT 3.5, consisting of 150 sentences—50\% with double negation and 50\% without. 
The sentences are hand-reviewed to ensure quality before evaluation. Table~\ref{TableDN} presents the results of the evaluated thresholds $n_t$ in terms of accuracy (ACC), precision (PRE), recall (REC), and F1 score (F1). 
Based on a comprehensive evaluation, we set $n_t$ to 3 and $w$ to 0.15.
When ChatGPT 3.5 was used to annotate the same dataset, it achieved an accuracy of 0.8333, precision of 1.000, recall of 0.6667, and an F1 score of 0.8. While highly effective at identifying detected double negatives, ChatGPT may mistakenly label litotes as not containing double negatives.
Litotes are not uncommon in English and involve using a double negative for effect. Although these are still considered double negatives, they may result in an alterfactual version of a text having the same meaning as the original, potentially confusing users.
Our algorithm achieves a higher F1 score, though at the cost of slightly reduced precision. Based on the results from Table~\ref{TableDN}, we set the threshold for determining double negatives to $n_t \leftarrow 0.20$.
We also present an ablation study in Table~\ref{TableDN_2} to evaluate Double Negative Detection with varying probability thresholds $n_t$ using $w = 3$.

\begin{table}[]
\centering
\footnotesize
\caption{\textbf{Double Negative Detection with Varying Window Size $w$ with $n_t$ = 0.15. The best and second best results are highlighted in bold and \underline{underline}.}}
\begin{tabular}{p{1cm} p{1cm} p{1cm} p{1cm} p{1cm}}
\hline
\textbf{$w$} & \textbf{ACC$\uparrow$} & \textbf{PRE$\uparrow$} & \textbf{REC$\uparrow$} & \textbf{F1$\uparrow$} \\ \hline
\hline
$1$ & 0.7987& 0.6133& \underline{0.9787}& 0.7541\\
$2$ & \textbf{0.9195}& \underline{0.8533}& \textbf{0.9846}& \underline{0.9143}\\
$3$ & \textbf{0.9195}& \textbf{0.8667}& 0.9702& \textbf{0.9155}\\
$4$ & \underline{0.9128}& \textbf{0.8667}& 0.9559& 0.9091\\
$5$ & 0.9060& \textbf{0.8667}& 0.9420& 0.9028\\
$6$ & 0.9060& \textbf{0.8667}& 0.9420& 0.9028\\
$7$ & 0.9060& \textbf{0.8667}& 0.9420& 0.9028\\
\hline
\end{tabular}
\label{TableDN}
\end{table}

\begin{table}[]
\centering
\footnotesize
\caption{\textbf{Double Negative Detection with Varying Probability Threshold $n_t$ with $w$ = 3. The best and second best results are highlighted in bold and \underline{underline}.}}
\begin{tabular}{p{1cm} p{1cm} p{1cm} p{1cm} p{1cm}}
\hline
\textbf{$n_t$} & \textbf{ACC$\uparrow$} & \textbf{PRE$\uparrow$} & \textbf{REC$\uparrow$} & \textbf{F1$\uparrow$} \\ \hline
\hline
$0.05$ & \textbf{0.9195}& \textbf{0.8667}& \textbf{0.9702}& \textbf{0.9155}\\
$0.1$ & \textbf{0.9195}& \textbf{0.8667}& \textbf{0.9702}& \textbf{0.9155}\\
$0.15$ & \textbf{0.9195}& \textbf{0.8667}& \textbf{0.9702}& \textbf{0.9155}\\
$0.20$ & \underline{0.9128}& \underline{0.8533}& \underline{0.9697}& \underline{0.9078}\\
$0.25$ & 0.8993& 0.8267& 0.9688& 0.8921\\
$0.30$ & 0.8792& 0.7867& 0.9672& 0.8675\\
$0.35$ & 0.8524& 0.7333& 0.9649& 0.8333\\
$0.40$ & 0.7987& 0.6267& 0.9592& 0.7581\\
\hline
\end{tabular}
\label{TableDN_2}
\end{table}

\renewcommand{\tabcolsep}{1.7pt}
\begin{table}[tb!]
\centering
\footnotesize
\begin{tabular}{cccccc}
\toprule
\textbf{Dataset} & \textbf{\#Avgwords} & \textbf{\#Labels} & \textbf{DistillBERT} & \textbf{BERT} & \textbf{RoBERTa}\\
\cmidrule(lr){1-6}
GB & 9.3 & 2 & 0.83 & 0.82 & 0.84 \\
HS & 13.72 & 2 & 0.67 & 0.97 & 0.98\\
EMO & 19.15 & 6 & 0.72 & 0.91 & 0.93 \\
JIG & 43.38 & 2 & 0.65 & 0.66 & 0.73\\
\bottomrule
\end{tabular}
\caption{Dataset statistics and accuracy of DistilBERT, BERT, and RoBERTa classification models on the test set.}
\label{TableModels}
\end{table}

\section{Detailed on Datasets Statistic}
Table~\ref{TableModels} presents detailed dataset statistics and the models used to evaluate {\mymethod}.

\section{Implementation Details}
We set our confidence threshold to $\sigma \leftarrow 0.05$ to ensure that the model's output confidence shifts by no more than 5\%. Sentence grammar similarity is maintained, as assessed by USE \citep{Cer2018-vh}, with a threshold of 0.8. Additionally, we constrain {\mymethod}'s perturbations by avoiding repeated perturbations and excluding a list of stopwords.
For ConceptNet, we set a minimum weight threshold of $\omega_t \leftarrow 0.5$ to ensure that the queried relations between words are sufficiently strong. For each dataset, we apply {\mymethod} in two ways: once with the task of perturbing only one word (denoted by '-Single' in the dataset name in Table~\ref{TableModels}), and once with the task of perturbing as many words as possible (denoted by '-Multi').

\section{Detail on Study Design}
\begin{figure*}[]
\centering      
\includegraphics[scale=0.7]{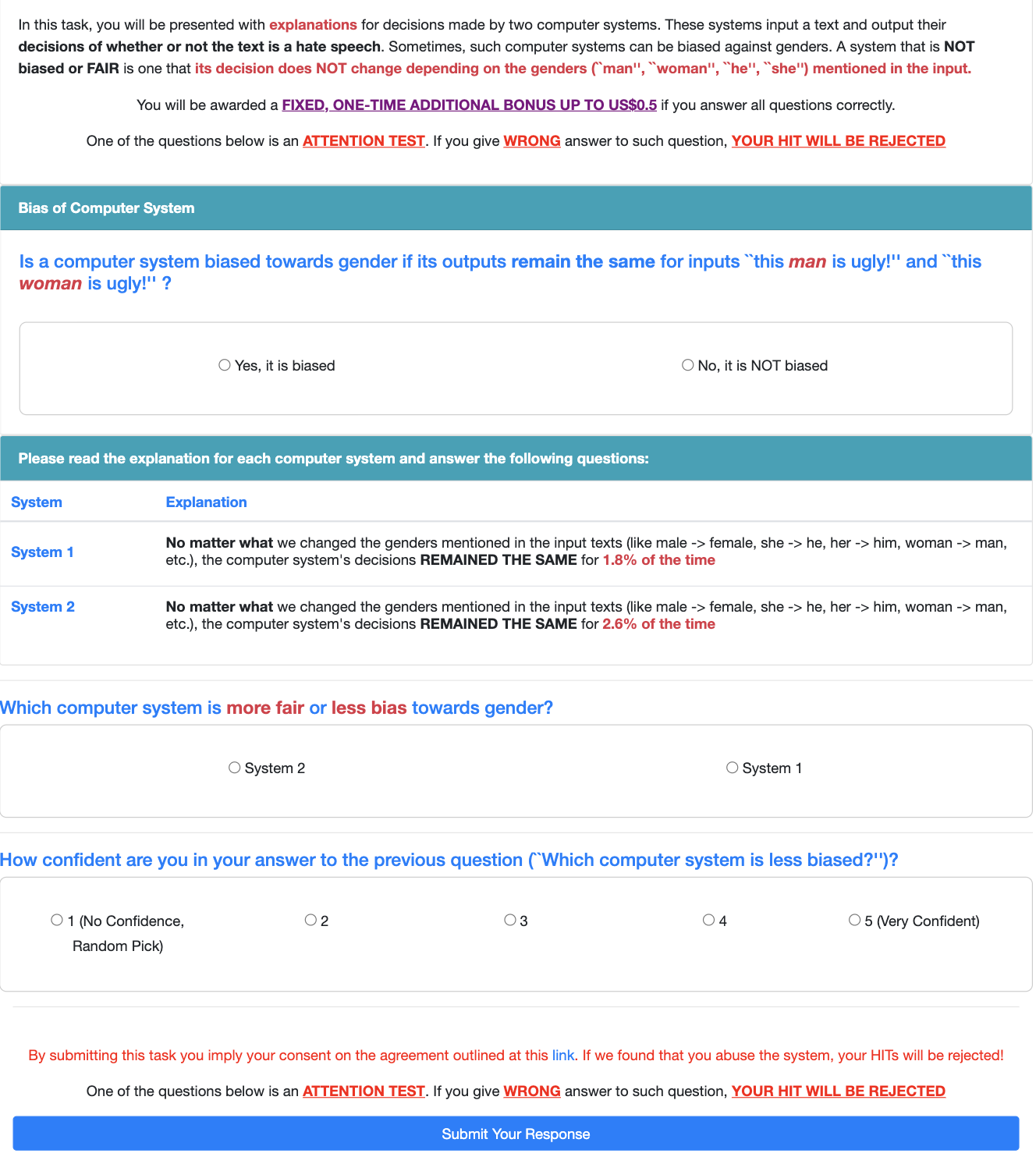}   
\caption{Front-end design is used in the user study.}
\label{fig:user_study3}
\end{figure*}

We provide a user study design template in Fig.~\ref{fig:user_study3}.

\end{document}